\documentclass[11pt]{article}
\usepackage[preprint]{acl}
\usepackage{times}
\usepackage{latexsym}

\usepackage[T1]{fontenc}

\usepackage[utf8]{inputenc}

\usepackage{microtype}

\usepackage{inconsolata}

\usepackage{graphicx}

\usepackage{booktabs}
\usepackage{multirow}
\usepackage{textcomp}
\title{Temporal Tokenization Strategies for Event Sequence Modeling with\\Large Language Models}

\author{
    Zefang Liu\textsuperscript{1},
    Nam H. Nguyen\textsuperscript{1},
    Yinzhu Quan\textsuperscript{2},
    Shi-Xiong Zhang\textsuperscript{1}
\\
\\
    \textsuperscript{1}Capital One, USA \\
    \textsuperscript{2}Georgia Institute of Technology, USA \\
\\
    \textbf{Correspondence:} \href{mailto:zefang.liu@capitalone.com}{zefang.liu@capitalone.com}
}
\begin{document}
\maketitle
\begin{abstract}
Representing continuous time is a critical and under-explored challenge in modeling temporal event sequences with large language models (LLMs). Various strategies like byte-level representations or calendar tokens have been proposed. However, the optimal approach remains unclear, especially given the diverse statistical distributions of real-world event data, which range from smooth log-normal to discrete, spiky patterns. This paper presents a systematic empirical study of temporal tokenization for modeling event sequences with LLMs, comparing distinct encoding strategies: naive numeric strings, high-precision byte-level representations, human-semantic calendar tokens, classic uniform binning, and adaptive residual scalar quantization. We evaluate these strategies by fine-tuning LLMs on real-world datasets that exemplify these diverse distributions. Our analysis reveals that no single strategy is universally superior; instead, prediction performance depends heavily on aligning the tokenizer with the data's statistical properties, highlighting temporal tokenization as a critical yet often overlooked design dimension in LLM-based event modeling.
\end{abstract}
%%%%%%%%%%%%%%%%%%%%%%%%%%%%%%%%%%%%%%%%
\section{Introduction}

Large language models (LLMs) have demonstrated remarkable capabilities in processing text, leading to new opportunities in modeling sequences of discrete events. While traditional temporal point processes (TPPs) \citep{hawkes2018hawkes,reinhart2018review,shchur2021neural,zhou2025advances} provide a framework, recent approaches leveraging LLMs \citep{liu2024tpp,kong2026byte} uniquely capture the rich textual semantics of events, framing event sequence modeling as a structured generative task that extends the boundaries of language-centric modeling. This integration, however, presents a fundamental challenge: LLMs are inherently discrete text processors, while real-world events occur in continuous time. The critical step of converting continuous time values into discrete tokens, known as temporal tokenization, is a crucial but under-explored component.

The optimal strategy for temporal tokenization is unclear. While the most naive approach is to represent time as a simple numeric string \citep{bhatia2025date}, other methods include human-semantic calendar tokens \citep{he2025efficient,he2025rhythm} and byte-level representations \citep{kong2026byte}. This choice is complicated by the diverse nature of real-world event data. Our analysis of several event sequence datasets reveals that temporal distributions vary substantially; they range from smooth log-normal patterns to discrete, spiky distributions and complex mixed-modal structures. A tokenization strategy that excels on one distribution may fail on another.

This paper presents a systematic empirical study of temporal tokenization strategies for event sequence modeling with LLMs\footnote{GitHub repository: \url{https://github.com/CapitalOne-Research/temporal-tokenization}}. We conduct a comprehensive comparison of five distinct encoding strategies: naive numeric strings, high-precision byte-level representations, human-semantic calendar tokens, classic uniform binning in linear or log space, and adaptive, data-driven residual scalar quantization (RSQ). By fine-tuning LLMs on a suite of real-world datasets that exemplify these diverse distributions, we analyze the resulting trade-offs between prediction performance and computational efficiency. Our findings establish a practical guide on data-tokenizer alignment, demonstrating that no single strategy is universally optimal, but that performance is maximized by aligning the tokenization scheme with the data’s specific statistical distribution.

%%%%%%%%%%%%%%%%%%%%%%%%%%%%%%%%%%%%%%%%
\section{Related Work}

Integrating LLMs with event sequence modeling presents a fundamental conflict in representing continuous time for discrete text processors. Two distinct strategies have emerged: hybrid frameworks that treat time as a vector embedding \citep{liu2024tpp,liu2025retrieval}, and pure tokenization approaches that convert continuous time intervals into discrete byte-level tokens \citep{kong2026byte}. This methodological divergence, intensified by the challenges posed by benchmarks like EasyTPP \citep{xue2024easytpp} and DanmakuTPPBench \citep{jiang2025danmakutppbench}, highlights that the need to solve this core representation problem remains unsettled. Research in adjacent domains confirms that varied tokenization strategies are critical for success: naive numeric strings have been shown to be problematic \citep{bhatia2025date}, while specialized methods like hierarchical calendar tokens \citep{he2025efficient,he2025rhythm} and patch-wise binning \citep{han2025patchcat} succeed in mobility and time series forecasting. Our paper addresses this gap by providing a systematic empirical comparison of these temporal tokenization strategies for event sequence modeling across diverse statistical distributions.

%%%%%%%%%%%%%%%%%%%%%%%%%%%%%%%%%%%%%%%%
\section{Temporal Tokenization}

An event sequence is a series $\mathcal{S} = \{(t_1, k_1), (t_2, k_2), \dots, (t_N, k_N)\}$, where each event $i$ consists of a continuous time value $t_i$ and a discrete text-based event type $k_i$. To model such sequences with an LLM, both components must be converted into discrete tokens. While the textual event type $k_i$ is naturally handled by the LLM's native tokenizer, the continuous time value $t_i$ requires a specialized temporal tokenization strategy to represent either the absolute time $t_i$ or, more commonly, the relative time interval $\Delta t_i = t_i - t_{i-1}$.

%%%%%%%%%%%%%%%%%%%%%%%%%%%%%%%%%%%%%%%%
\subsection{Numeric String Tokenization}

The naive numeric string strategy formats the continuous time value $v_i$ (e.g., the time interval) into a plain text string with a fixed precision, such as ``0.076''. This string is then passed directly to the LLM's standard subword tokenizer, a process that often fragments it into semantically meaningless subwords (e.g., ``0'', ``.'', ``07'', ``6''). The approach requires no modifications to the model's vocabulary, instead relying entirely on the LLM's pretrained ability to parse and reason about numerical strings as standard text.

%%%%%%%%%%%%%%%%%%%%%%%%%%%%%%%%%%%%%%%%
\subsection{Byte Tokenization}

Byte tokenization, a strategy introduced by Language-TPP \citep{kong2026byte}, bypasses the LLM's subword tokenizer to enable seamless integration of continuous time. The approach first represents the time value $v_i$ as a standard 32-bit float, then deterministically splits its 4-byte representation into four discrete byte values. The LLM's vocabulary is augmented with 256 new special tokens (e.g., \texttt{<|byte\_0|>} \dots \texttt{<|byte\_255|>}) to represent each possible byte value. The float $v_i$ is thus always encoded as a fixed-length sequence of four tokens, preserving full float32 precision in a token-efficient manner.

%%%%%%%%%%%%%%%%%%%%%%%%%%%%%%%%%%%%%%%%
\subsection{Calendar Tokenization}

The calendar tokenization strategy is human-centric, converting a time value into a sequence of semantically meaningful tokens based on the Gregorian calendar. An absolute timestamp $t_i$ can be encoded into tokens like \texttt{<|year\_2025|>}, \texttt{<|month\_10|>}, and \texttt{<|day\_25|>}. Similarly, a relative interval $\Delta t_i$ can be encoded into tokens such as \texttt{<|days\_03|>} and \texttt{<|hours\_05|>}. This approach requires adding calendar-based tokens to the vocabulary and relies on the LLM's pretrained world knowledge of calendar systems.

%%%%%%%%%%%%%%%%%%%%%%%%%%%%%%%%%%%%%%%%
\subsection{Scale Bin Tokenization}

Scale bin tokenization represents a classic TPP-based uniform binning approach. It first transforms all time values $v_i$ into a target space (either linear or log scale). During training, the strategy finds the minimum and maximum of these transformed values and divides this range into $K$ (e.g., 256) uniform bins. Each bin is then assigned a unique token (e.g., \texttt{<|bin\_000|>} \dots \texttt{<|bin\_255|>}) added to the vocabulary. During decoding, the continuous time value is reconstructed by mapping the token back to the center of its corresponding bin in the transformed space. This method is data-driven but assumes a uniform distribution within the transformed space.

%%%%%%%%%%%%%%%%%%%%%%%%%%%%%%%%%%%%%%%%
\subsection{Residual Scalar Quantization Tokenization}

Residual scalar quantization (RSQ) tokenization is an adaptive multi-stage binning strategy analogous to residual vector quantization (RVQ) \citep{gray1984vector,makhoul1985vector}. This method first transforms all time values $v_i$ into a target space (e.g., linear or log scale), yielding $v_i' = f(v_i)$. Instead of using uniform bins, it employs a hierarchy of quantizers. At the first level, a K-Means model is trained on the transformed data $\{v_i'\}$ to find an initial codebook of $K_1$ cluster centers, $C^{(1)} = \{c_1^{(1)}, \dots, c_{K_1}^{(1)}\}$. A value $v_i'$ is quantized to its nearest centroid $c_{q_1}^{(1)}$, where $q_1$ is the token index for the first level. The process then becomes iterative: the residual error from this first quantization, $r_1 = v_i' - c_{q_1}^{(1)}$, is calculated and passed to the second level. A second K-Means model is trained on these first-level residuals $\{r_1\}$ to find a new codebook $C^{(2)}$ with $K_2$ centers. This second level quantizes $r_1$ to its nearest centroid $c_{q_2}^{(2)}$, yielding the second token $q_2$ and a new residual $r_2 = r_1 - c_{q_2}^{(2)}$. This is repeated for $N$ levels, encoding the single scalar $v_i'$ as a fixed-length sequence of $N$ discrete token indices $\langle q_1, q_2, \dots, q_N \rangle$. These are mapped to unique tokens (e.g., \texttt{<|L0\_015|>}, \texttt{<|L1\_122|>}, \dots), creating a fine-grained, compositional representation. Decoding is a simple additive process where the reconstructed value $\hat{v}_i'$ is the sum of the centroids from each level's codebook, $\hat{v}_i' = c_{q_1}^{(1)} + c_{q_2}^{(2)} + \dots + c_{q_N}^{(N)}$, before applying the inverse transform $v_i = f^{-1}(\hat{v}_i')$.

%%%%%%%%%%%%%%%%%%%%%%%%%%%%%%%%%%%%%%%%
\section{Experiments}

In this section, we empirically evaluate the temporal tokenization strategies discussed previously.

%%%%%%%%%%%%%%%%%%%%%%%%%%%%%%%%%%%%%%%%
\subsection{Data}

\begin{figure}[h!]
\centering
\includegraphics[width=\columnwidth]{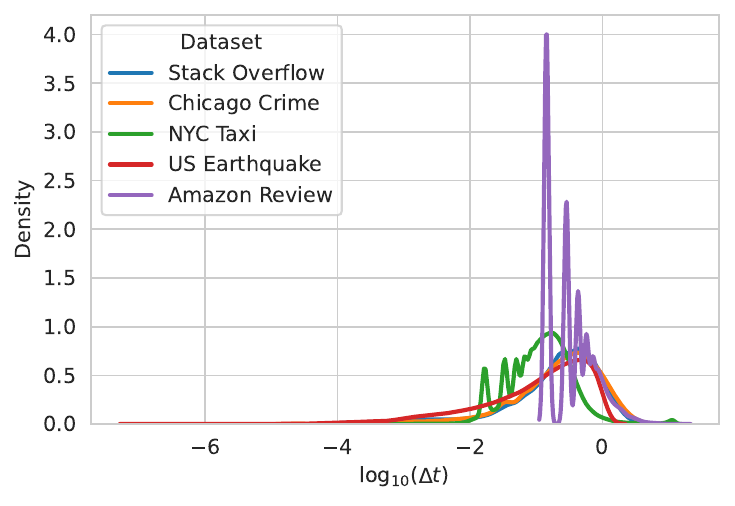}
\caption{Log-scale distributions of relative time intervals ($\Delta t_i$) across all five datasets.}
\label{fig:dist-log}
\end{figure}

We conduct our experiments using the five real-world datasets from TPP-LLM\footnote{Datasets: \url{https://huggingface.co/tppllm}} \citep{liu2024tpp}. As detailed in Table~\ref{tab:dataset-stats}, these datasets provide sequences of textual event types, raw absolute timestamps, and pre-computed relative time intervals with proper units. Calendar-based tokenizers ingest the absolute timestamps, while the numeric, byte, and binning-based strategies operate on the relative time intervals. We also analyze the statistical properties of the relative time intervals ($\Delta t_i$) from each dataset. As shown in Figure~\ref{fig:dist-log}, the distributions are highly diverse. They range from smooth, log-normal patterns (e.g., Stack Overflow) and multi-modal curves (e.g., NYC Taxi) to extremely discrete, spiky distributions with sharp peaks at specific intervals (e.g., Amazon Review). More dataset distributions and examples can be found in Appendix \ref{app:data-distributions} and \ref{app:data-examples}.

\begin{table}[h!]
\centering
\caption{Dataset statistics, including the number of event types, total events, sequences with their train/validation/test splits, average sequence length, and the temporal unit for intervals.}
\label{tab:dataset-stats}
\resizebox{\linewidth}{!}{%
\begin{tabular}{lrrrrrl}
\toprule
\textbf{Dataset} & \textbf{Types} & \textbf{Events} & \textbf{Seqs} & \textbf{Train/Val/Test} & \textbf{Seq Len} & \textbf{Unit} \\
\midrule
Stack Overflow & 25 & 187,836 & 3,336 & 2,668/334/334 & 56.31 & Month \\
Chicago Crime & 20 & 202,867 & 4,048 & 3,238/405/405 & 50.12 & Month \\
NYC Taxi & 8 & 362,370 & 2,957 & 2,365/296/296 & 122.55 & Hour \\
US Earthquake & 3 & 30,218 & 3,079 & 2,463/308/308 & 9.81 & Day \\
Amazon Review & 18 & 127,054 & 2,245 & 1,796/224/225 & 56.59 & Week \\
\bottomrule
\end{tabular}
}
\end{table}

%%%%%%%%%%%%%%%%%%%%%%%%%%%%%%%%%%%%%%%%
\begin{table*}[h!]
\centering
\caption{Main experimental results comparing temporal tokenization strategies. Next event type accuracy (Acc \% $\uparrow$) and next event time RMSE ($\downarrow$) are reported across all five datasets. Tokens ($\downarrow$) measures the number of tokens required to represent a single time value. Results are averaged over 5 runs, with standard deviations reported as subscripts. (Abs. = Absolute; Rel. = Relative; Sec. = Second; L1/L4 = 1 or 4 levels.)}
\label{tab:main-results}
\resizebox{\textwidth}{!}{%
\setlength{\tabcolsep}{3pt}
\begin{tabular}{l c rr rr rr rr rr}
\toprule
& & \multicolumn{2}{c}{\textbf{Stack Overflow}} & \multicolumn{2}{c}{\textbf{Chicago Crime}} & \multicolumn{2}{c}{\textbf{NYC Taxi}} & \multicolumn{2}{c}{\textbf{US Earthquake}} & \multicolumn{2}{c}{\textbf{Amazon Review}} \\
\cmidrule(lr){3-4} \cmidrule(lr){5-6} \cmidrule(lr){7-8} \cmidrule(lr){9-10} \cmidrule(lr){11-12}
\textbf{Tokenizer} & \textbf{Tokens} & \textbf{Acc $\uparrow$} & \textbf{RMSE $\downarrow$} & \textbf{Acc $\uparrow$} & \textbf{RMSE $\downarrow$} & \textbf{Acc $\uparrow$} & \textbf{RMSE $\downarrow$} & \textbf{Acc $\uparrow$} & \textbf{RMSE $\downarrow$} & \textbf{Acc $\uparrow$} & \textbf{RMSE $\downarrow$} \\
\midrule
% --- Strategy 1: Naive String ---
Numeric String & \textasciitilde4 & 44.5$_{\pm.1}$ & 0.625$_{\pm.000}$ & \underline{27.1}$_{\pm.1}$ & 0.728$_{\pm.000}$ & \underline{92.0}$_{\pm.0}$ & \underline{0.888}$_{\pm.002}$ & 64.0$_{\pm.2}$ & 0.381$_{\pm.000}$ & \underline{69.9}$_{\pm.1}$ & 0.647$_{\pm.000}$ \\
\midrule
% --- Strategy 2: Byte-level ---
Byte & 4 & 44.3$_{\pm.0}$ & 0.516$_{\pm.004}$ & \textbf{27.2}$_{\pm.1}$ & 0.726$_{\pm.000}$ & \underline{92.0}$_{\pm.0}$ & 0.892$_{\pm.001}$ & 64.0$_{\pm.1}$ & \textbf{0.348}$_{\pm.006}$ & 69.7$_{\pm.0}$ & 0.644$_{\pm.000}$ \\
\midrule
% --- Strategy 3: Human-centric ---
Abs. Calendar (Day) & 3 & \underline{44.7}$_{\pm.1}$ & 0.578$_{\pm.002}$ & \underline{27.1}$_{\pm.1}$ & 0.698$_{\pm.001}$ & 91.9$_{\pm.1}$ & 16.086$_{\pm.009}$ & \textbf{64.4}$_{\pm.3}$ & 0.838$_{\pm.002}$ & \textbf{70.0}$_{\pm.1}$ & 0.633$_{\pm.002}$ \\
Abs. Calendar (Sec.) & 6 & \textbf{44.8}$_{\pm.1}$ & 0.569$_{\pm.003}$ & \underline{27.1}$_{\pm.1}$ & 0.692$_{\pm.001}$ & \textbf{92.2}$_{\pm.0}$ & \textbf{0.869}$_{\pm.006}$ & 64.2$_{\pm.3}$ & \underline{0.353}$_{\pm.001}$ & \textbf{70.0}$_{\pm.1}$ & 0.634$_{\pm.001}$ \\
Rel. Calendar (Day) & 3 & 44.4$_{\pm.1}$ & 0.626$_{\pm.000}$ & \underline{27.1}$_{\pm.1}$ & 0.728$_{\pm.000}$ & 91.9$_{\pm.0}$ & 0.912$_{\pm.000}$ & 64.0$_{\pm.2}$ & 0.381$_{\pm.000}$ & 69.8$_{\pm.1}$ & 0.648$_{\pm.001}$ \\
Rel. Calendar (Sec.) & 6 & 44.5$_{\pm.1}$ & 0.623$_{\pm.001}$ & \textbf{27.2}$_{\pm.1}$ & 0.728$_{\pm.000}$ & \underline{92.0}$_{\pm.0}$ & 0.891$_{\pm.002}$ & \underline{64.3}$_{\pm.2}$ & 0.381$_{\pm.000}$ & 69.7$_{\pm.1}$ & 0.648$_{\pm.000}$ \\
\midrule
% --- Strategy 4: Uniform Binning ---
Scale Bin (Linear) & 1 & 44.3$_{\pm.1}$ & 0.615$_{\pm.000}$ & \underline{27.1}$_{\pm.1}$ & 0.717$_{\pm.000}$ & \underline{92.0}$_{\pm.0}$ & 0.894$_{\pm.007}$ & 64.0$_{\pm.3}$ & 0.376$_{\pm.004}$ & 69.7$_{\pm.1}$ & 0.637$_{\pm.001}$ \\
Scale Bin (Log) & 1 & 44.3$_{\pm.0}$ & \underline{0.479}$_{\pm.003}$ & \underline{27.1}$_{\pm.1}$ & \textbf{0.562}$_{\pm.004}$ & \underline{92.0}$_{\pm.0}$ & 1.127$_{\pm.061}$ & 63.9$_{\pm.2}$ & 0.360$_{\pm.015}$ & 69.6$_{\pm.1}$ & \underline{0.632}$_{\pm.001}$ \\
\midrule
% --- Strategy 5: Adaptive Binning (RSQ) ---
RSQ (Linear, L1) & 1 & 44.2$_{\pm.1}$ & 0.595$_{\pm.004}$ & \underline{27.1}$_{\pm.1}$ & 0.708$_{\pm.003}$ & \underline{92.0}$_{\pm.0}$ & 0.911$_{\pm.017}$ & 63.7$_{\pm.3}$ & 0.355$_{\pm.003}$ & 69.7$_{\pm.2}$ & 0.647$_{\pm.001}$ \\
RSQ (Linear, L4) & 4 & 44.5$_{\pm.0}$ & 0.603$_{\pm.002}$ & \underline{27.1}$_{\pm.1}$ & 0.698$_{\pm.003}$ & \underline{92.0}$_{\pm.0}$ & 0.891$_{\pm.007}$ & 64.1$_{\pm.2}$ & 0.363$_{\pm.002}$ & 69.7$_{\pm.1}$ & 0.646$_{\pm.001}$ \\
RSQ (Log, L1) & 1 & 44.3$_{\pm.1}$ & 0.497$_{\pm.005}$ & \underline{27.1}$_{\pm.2}$ & 0.576$_{\pm.009}$ & \underline{92.0}$_{\pm.0}$ & 1.221$_{\pm.040}$ & 63.8$_{\pm.2}$ & 0.368$_{\pm.010}$ & 69.6$_{\pm.1}$ & \underline{0.632}$_{\pm.001}$ \\
RSQ (Log, L4) & 4 & 44.2$_{\pm.1}$ & \textbf{0.474}$_{\pm.003}$ & \textbf{27.2}$_{\pm.1}$ & \underline{0.575}$_{\pm.003}$ & \underline{92.0}$_{\pm.0}$ & 1.201$_{\pm.057}$ & 64.2$_{\pm.2}$ & 0.366$_{\pm.010}$ & 69.7$_{\pm.1}$ & \textbf{0.631}$_{\pm.000}$ \\
\midrule
\midrule
TPP-LLM\footnotemark[1] & - & 44.4$_{{\pm.1}}$ & 0.464$_{{\pm.001}}$ & 27.3$_{{\pm.1}}$ & 0.542$_{{\pm.001}}$ & 91.8$_{{\pm.1}}$ & 0.826$_{{\pm.002}}$ & 64.1$_{{\pm.3}}$ & 0.265$_{{\pm.001}}$ & 69.5$_{{\pm.1}}$ & 0.570$_{{\pm.002}}$ \\
\bottomrule
\end{tabular}
}
\end{table*}
%%%%%%%%%%%%%%%%%%%%%%%%%%%%%%%%%%%%%%%%
\subsection{Experimental Setup}

We fine-tune a pretrained LLM backbone Llama-3.2-1B \citep{grattafiori2024llama} with QLoRA \citep{dettmers2023qlora} through the Hugging Face framework \citep{wolf2020transformers} for all experiments and use TPP-LLM \citep{liu2024tpp} as the baseline. We format each event using a consistent template inspired from Language-TPP \citep{kong2026byte}: \texttt{<|begin\_of\_event|>}\allowbreak\texttt{<|type\_prefix|>}\allowbreak\texttt{\{type\_tokens\}}\allowbreak\texttt{<|time\_prefix|>}\allowbreak\texttt{\{time\_tokens\}}\allowbreak\texttt{<|end\_of\_event|>}. We compare multiple temporal tokenization strategies, including numeric strings (6-decimal precision), byte-level tokens, calendar tokens (at second and day resolution), uniform scale binning (256 bins, linear and base-10 logarithm scale), and multi-level RSQ (1 level of 256 bins and 4 levels of 64 bins each, linear and base-10 logarithm scale).

We evaluate model performance on prediction quality, measured by the accuracy of the next event type prediction and the root mean squared error (RMSE) of the predicted time interval, and efficiency, reported by the number of temporal tokens to represent a single time value for each strategy, serving as a hardware-agnostic proxy for inference latency. For strategies predicting absolute timestamps, the RMSE is computed directly between the predicted and true timestamps (inherently reflecting the relative interval error), with all final RMSEs normalized by the dataset-specific temporal units detailed in Table~\ref{tab:dataset-stats}. During inference, if a tokenizer generates malformed or partial sequences that fail to parse, the predicted time interval defaults to zero to ensure a robust error penalty during evaluation. We run 5 independent experiments with different random seeds for each setting, reporting the averaged scores alongside their standard deviations. Additional training and evaluation details are provided in Appendix \ref{app:experimental-settings}.

%%%%%%%%%%%%%%%%%%%%%%%%%%%%%%%%%%%%%%%%
\subsection{Experimental Results}

Our main findings, presented in Table~\ref{tab:main-results}, reveal a clear and important divergence: the choice of temporal tokenizer has a minimal impact on event type prediction but a significant and data-dependent impact on event time prediction.

\footnotetext[1]{TPP-LLM results are included for indirect comparison, as it uses a distinct framework employing TPP-specific prediction heads and a TPP-based loss function.}

For the next event type accuracy, all strategies perform similarly across all five datasets. Conversely, the performance on next event time (RMSE) shows dramatic variation, supporting the observation that a tokenizer must align with the dataset's underlying temporal distribution. For the datasets with skewed log-normal distributions (Stack Overflow, Chicago Crime) and the spiky log-scale distribution (Amazon Review), the log-based strategies consistently outperform others. Scale Bin (Log) achieves the best RMSE on Chicago Crime, while RSQ (Log, L4) is the top performer on Stack Overflow and Amazon Review. Intuitively, log-based tokenization allocates a higher density of discrete bins to smaller time intervals, preserving critical precision in the dense regions of these skewed distributions while efficiently compressing the long tail. However, these same log-based strategies perform poorly on the mixed distribution of NYC Taxi, where Absolute Calendar (Second) is the most effective. Unlike the other datasets, NYC Taxi exhibits strong daily and hourly cyclic patterns dictated by human mobility; calendar tokens explicitly capture this intrinsic periodicity, whereas simple interval binning obscures it. For the US Earthquake dataset, which also exhibits a skewed log-normal distribution, Byte achieves the lowest error, suggesting that its high-precision floating-point representation is advantageous for strictly continuous scientific measurements.

This outcome demonstrates that the optimal tokenization strategy is highly dependent on the data's specific statistical properties, and a similar trade-off applies to efficiency. The single-token log-scale binning and RSQ strategies provide an excellent balance, achieving top-tier RMSE performance on several datasets while being the most token-efficient. In contrast, multi-token strategies like byte and multi-level RSQ explicitly trade token efficiency for higher precision.

While our pure LLM-based approach results in higher RMSEs on several datasets compared to TPP-LLM \citep{liu2024tpp}, it offers a simpler and more scalable architecture. Our method treats time prediction as a standard next-token prediction task, entirely avoiding specialized TPP prediction heads, the complex TPP log-likelihood loss function, and substantial computational burden from its integral term. This architectural simplicity, which still maintains competitive event type accuracy, streamlines the modeling process and makes large-scale continual pre-training on event sequences a more feasible and promising direction for future work. More ablation studies can be found in Appendix \ref{app:ablation-studies}.

%%%%%%%%%%%%%%%%%%%%%%%%%%%%%%%%%%%%%%%%
\section{Conclusion}

We presented a systematic empirical study comparing temporal tokenization strategies for modeling event sequences with LLMs. Our findings reveal a critical trade-off: while event type prediction accuracy is largely insensitive to the choice of time tokenizer, next event time prediction is highly dependent on the alignment between the tokenization strategy and the dataset's underlying statistical distribution. We found that log-based strategies (RSQ and scale binning) excel on datasets with log-normal or spiky-log distributions, while calendar tokenization approaches are more robust for mixed-modal distributions. Our work provides a practical guide for this crucial design choice and validates that a pure LLM fine-tuning framework is a simple yet competitive alternative to complex hybrid temporal point process models.

%%%%%%%%%%%%%%%%%%%%%%%%%%%%%%%%%%%%%%%%
\section*{Limitations}

Our study, while systematic, has several limitations. The experiments were conducted primarily on Llama-3.2 models up to 3B parameters; these findings may not fully generalize to larger models or different model families. While our five datasets were chosen for their diverse distributions, they do not represent the entire spectrum of real-world event data. Our evaluation is also focused on next-event prediction, and these trade-offs might differ for other tasks, such as long-horizon forecasting or sequence-level generation. Furthermore, the use of PEFT may introduce a representation bottleneck compared to full fine-tuning, potentially limiting the model's ability to fully integrate specialized temporal tokens into its global attention space. Finally, while we explored a representative set of temporal tokenizers and their configurations, an exhaustive search across the diverse landscape of tokenization algorithms, binning granularities, and quantization hierarchies remains an area for future investigation.

%%%%%%%%%%%%%%%%%%%%%%%%%%%%%%%%%%%%%%%%
\section*{Ethical Considerations}

Our work focuses on the technical trade-offs of tokenization, utilizing established, publicly available benchmarks that are anonymized and contain no personally identifiable information. However, because these datasets model human behaviors, improved predictive accuracy carries potential risks inherited from downstream applications. Crucially, our finding that temporal tokenization strategies are not universal indicates that applying a strategy optimized for one pattern to a mismatched domain may cause performance degradation and result in inconsistent predictive accuracy. Therefore, aligning the temporal tokenization strategy with the underlying data distribution is essential to ensure robust and reliable model performance.

%%%%%%%%%%%%%%%%%%%%%%%%%%%%%%%%%%%%%%%%
\bibliography{refs}

@article{hawkes2018hawkes,
  title={{Hawkes} Processes and Their Applications to Finance: A Review},
  author={Hawkes, Alan G.},
  journal={Quantitative Finance},
  volume={18},
  number={2},
  pages={193--198},
  year={2018},
  publisher={Taylor \& Francis},
  doi={10.1080/14697688.2017.1403131},
}

@article{reinhart2018review,
  title={A Review of Self-Exciting Spatio-Temporal Point Processes and Their Applications},
  author={Reinhart, Alex},
  journal={Statistical Science},
  volume={33},
  number={3},
  pages={299--318},
  year={2018},
  publisher={JSTOR},
  doi={10.1214/17-STS629},
}

@inproceedings{shchur2021neural,
  title={Neural Temporal Point Processes: A Review},
  author={Shchur, Oleksandr and T{\"u}rkmen, Ali Caner and Januschowski, Tim and G{\"u}nnemann, Stephan},
  booktitle={Proceedings of the Thirtieth International Joint Conference on Artificial Intelligence},
  pages={4585--4593},
  year={2021},
  organization={International Joint Conferences on Artificial Intelligence Organization},
  doi={10.24963/ijcai.2021/623},
}

@article{zhou2025advances,
  title={Advances in Temporal Point Processes: {Bayesian}, Neural, and {LLM} Approaches},
  author={Zhou, Feng and Kong, Quyu and Qiao, Jie and Wan, Cheng and Zhang, Yixuan and Cai, Ruichu},
  journal={arXiv preprint arXiv:2501.14291},
  year={2025},
  doi={10.48550/arXiv.2501.14291},
}

@article{liu2024tpp,
  title={{TPP-LLM}: Modeling Temporal Point Processes by Efficiently Fine-Tuning Large Language Models},
  author={Liu, Zefang and Quan, Yinzhu},
  journal={arXiv preprint arXiv:2410.02062},
  year={2024},
  doi={10.48550/arXiv.2410.02062},
}

@inproceedings{kong2026byte,
  title={Byte-Token Enhanced Language Models for Temporal Point Processes Analysis},
  author={Kong, Quyu and Zhang, Yixuan and Liu, Yang and Tong, Panrong and Liu, Enqi and Zhou, Feng},
  booktitle={Proceedings of the ACM Web Conference 2026},
  pages={7013--7023},
  year={2026},
  doi={10.1145/3774904.3792197},
}

@inproceedings{jiang2025danmakutppbench,
  title={{DanmakuTPPBench}: A Multi-modal Benchmark for Temporal Point Process Modeling and Understanding},
  author={Jiang, Yue and Li, Jichu and Liu, Yang and Yang, Dingkang and Zhou, Feng and Kong, Quyu},
  booktitle={Advances in Neural Information Processing Systems (NeurIPS)},
  volume={38},
  year={2025},
  publisher={Curran Associates, Inc.},
  url={https://proceedings.neurips.cc/paper_files/paper/2025/file/272564c77318dae969b58090e3df5520-Paper-Datasets_and_Benchmarks_Track.pdf},
}

@inproceedings{liu2025retrieval,
  title={Retrieval of Temporal Event Sequences from Textual Descriptions},
  author={Liu, Zefang and Quan, Yinzhu},
  booktitle={Proceedings of the 4th International Workshop on Knowledge-Augmented Methods for Natural Language Processing (KnowledgeNLP)},
  pages={37--49},
  year={2025},
  address={Albuquerque, New Mexico, USA},
  publisher={Association for Computational Linguistics},
  doi={10.18653/v1/2025.knowledgenlp-1.3},
}

@inproceedings{bhatia2025date,
  title={Date Fragments: A Hidden Bottleneck of Tokenization for Temporal Reasoning},
  author={Bhatia, Gagan and Peyrard, Maxime and Zhao, Wei},
  booktitle={Proceedings of the 2025 Conference on Empirical Methods in Natural Language Processing (EMNLP)},
  pages={3201--3219},
  year={2025},
  address={Suzhou, China},
  publisher={Association for Computational Linguistics},
  doi={10.18653/v1/2025.emnlp-main.159},
}

@misc{han2025patchcat,
  title={{PatchCat}: Rethinking Temporal Tokenization in Time Series Forecasting},
  author={Han, Xiao and Zhang, Xinfeng and Wu, Yiling and Zhang, Zhenduo and Deng, Zhiyuan and Wu, Zhe},
  year={2025},
  howpublished={ICLR 2026 Conference Withdrawn Submission},
  url={https://openreview.net/forum?id=vG9vqUwjbu},
}

@article{he2025efficient,
  title={Efficient Temporal Tokenization for Mobility Prediction with Large Language Models},
  author={He, Haoyu and Luo, Haozheng and Chen, Yan and Wang, Qi R.},
  journal={arXiv preprint arXiv:2507.14017},
  year={2025},
  doi={10.48550/arXiv.2507.14017},
}

@inproceedings{he2025rhythm,
  title={{RHYTHM}: Reasoning with Hierarchical Temporal Tokenization for Human Mobility},
  author={He, Haoyu and Luo, Haozheng and Chen, Yan and Wang, Qi R.},
  booktitle={Advances in Neural Information Processing Systems (NeurIPS)},
  volume={38},
  pages={84700--84730},
  year={2025},
  publisher={Curran Associates, Inc.},
  url={https://proceedings.neurips.cc/paper_files/paper/2025/file/7a3497d601db1801a4a1af8113b3684f-Paper-Conference.pdf},
}

@inproceedings{dettmers2023qlora,
  title={{QLoRA}: Efficient Finetuning of Quantized {LLMs}},
  author={Dettmers, Tim and Pagnoni, Artidoro and Holtzman, Ari and Zettlemoyer, Luke},
  booktitle={Advances in Neural Information Processing Systems (NeurIPS)},
  volume={36},
  pages={10088--10115},
  year={2023},
  url={https://proceedings.neurips.cc/paper_files/paper/2023/file/1feb87871436031bdc0f2beaa62a049b-Paper-Conference.pdf},
}

@inproceedings{hu2022lora,
  title={{LoRA}: Low-Rank Adaptation of Large Language Models},
  author={Hu, Edward J. and Shen, Yelong and Wallis, Phillip and Allen-Zhu, Zeyuan and Li, Yuanzhi and Wang, Shean and Wang, Lu and Chen, Weizhu},
  booktitle={International Conference on Learning Representations (ICLR)},
  year={2022},
  doi={10.48550/arXiv.2106.09685},
}

@article{grattafiori2024llama,
  title={The {Llama} 3 Herd of Models},
  author={{Llama Team}},
  journal={arXiv preprint arXiv:2407.21783},
  year={2024},
  doi={10.48550/arXiv.2407.21783},
}

@inproceedings{wolf2020transformers,
  title={{Transformers}: State-of-the-Art Natural Language Processing},
  author={Wolf, Thomas and Debut, Lysandre and Sanh, Victor and Chaumond, Julien and Delangue, Clement and Moi, Anthony and Cistac, Pierric and Rault, Tim and Louf, R{\'e}mi and Funtowicz, Morgan and Davison, Joe and Shleifer, Sam and von Platen, Patrick and Ma, Clara and Jernite, Yacine and Plu, Julien and Xu, Canwen and Le Scao, Teven and Gugger, Sylvain and Drame, Mariama and Lhoest, Quentin and Rush, Alexander},
  booktitle={Proceedings of the 2020 Conference on Empirical Methods in Natural Language Processing: System Demonstrations (EMNLP)},
  pages={38--45},
  year={2020},
  doi={10.18653/v1/2020.emnlp-demos.6},
}

@inproceedings{loshchilov2019decoupled,
  title={Decoupled Weight Decay Regularization},
  author={Loshchilov, Ilya and Hutter, Frank},
  booktitle={International Conference on Learning Representations (ICLR)},
  year={2019},
  doi={10.48550/arXiv.1711.05101},
}

@inproceedings{xue2024easytpp,
  title={{EasyTPP}: Towards Open Benchmarking Temporal Point Processes},
  author={Xue, Siqiao and Shi, Xiaoming and Chu, Zhixuan and Wang, Yan and Hao, Hongyan and Zhou, Fan and Jiang, Caigao and Pan, Chen and Zhang, James Y. and Wen, Qingsong and Zhou, Jun and Mei, Hongyuan},
  booktitle={International Conference on Learning Representations (ICLR)},
  year={2024},
  doi={10.48550/arXiv.2307.08097},
}

@article{gray1984vector,
  title={Vector Quantization},
  author={Gray, Robert},
  journal={IEEE ASSP Magazine},
  volume={1},
  number={2},
  pages={4--29},
  year={1984},
  publisher={IEEE},
  doi={10.1109/massp.1984.1162229},
}

@article{makhoul1985vector,
  title={Vector Quantization in Speech Coding},
  author={Makhoul, John and Roucos, Salim and Gish, Herbert},
  journal={Proceedings of the IEEE},
  volume={73},
  number={11},
  pages={1551--1588},
  year={1985},
  publisher={IEEE},
  doi={10.1109/PROC.1985.13340},
}

@article{pedregosa2011scikit,
  title={{Scikit-learn}: Machine Learning in {Python}},
  author={Pedregosa, Fabian and Varoquaux, Ga{\"e}l and Gramfort, Alexandre and Michel, Vincent and Thirion, Bertrand and Grisel, Olivier and Blondel, Mathieu and Prettenhofer, Peter and Weiss, Ron and Dubourg, Vincent and Vanderplas, Jake and Passos, Alexandre and Cournapeau, David and Brucher, Matthieu and Perrot, Matthieu and Duchesnay, {\'E}douard},
  journal={Journal of Machine Learning Research (JMLR)},
  volume={12},
  pages={2825--2830},
  year={2011},
  publisher={JMLR.org},
  url={https://www.jmlr.org/papers/v12/pedregosa11a.html},
}
%%%%%%%%%%%%%%%%%%%%%%%%%%%%%%%%%%%%%%%%
\appendix
%%%%%%%%%%%%%%%%%%%%%%%%%%%%%%%%%%%%%%%%
\section{Data Distributions}
\label{app:data-distributions}

\begin{figure}[h!]
    \centering
    \includegraphics[width=.9\columnwidth]{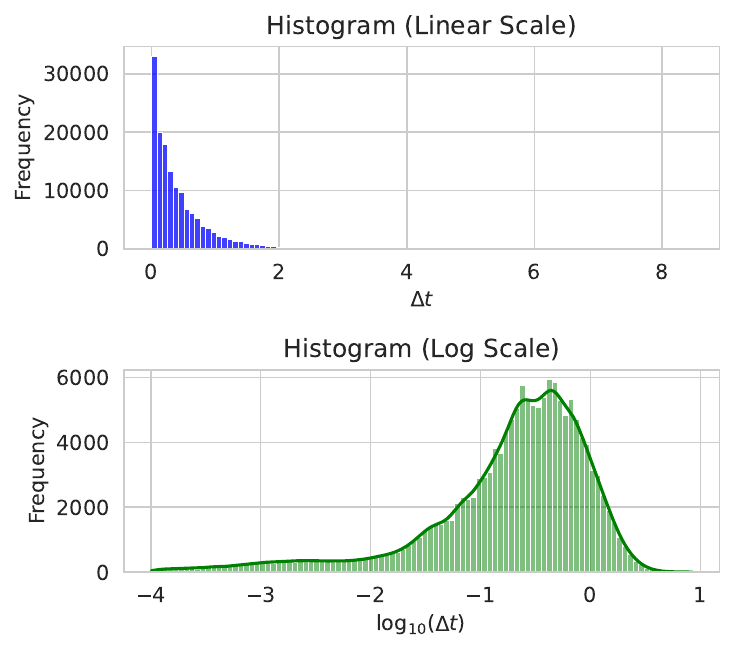}
    \caption{Distribution of relative time intervals ($\Delta t_i$) for the Stack Overflow dataset, showing linear scale (top) and log scale (bottom).}
    \label{fig:dist-stackoverflow}
\end{figure}

\begin{figure}[h!]
    \centering
    \includegraphics[width=.9\columnwidth]{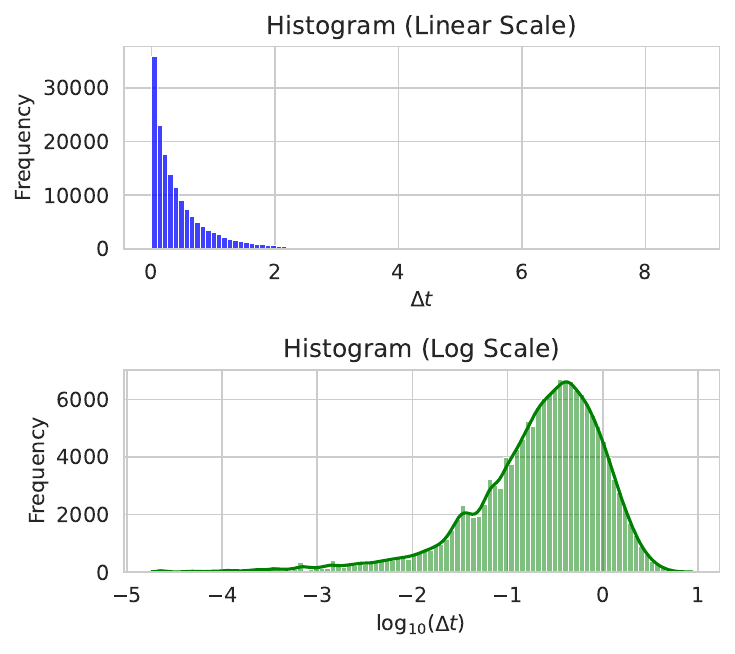}
    \caption{Distribution of relative time intervals ($\Delta t_i$) for the Chicago Crime dataset, showing linear scale (top) and log scale (bottom).}
    \label{fig:dist-chicago}
\end{figure}

\begin{figure}[h!]
    \centering
    \includegraphics[width=.9\columnwidth]{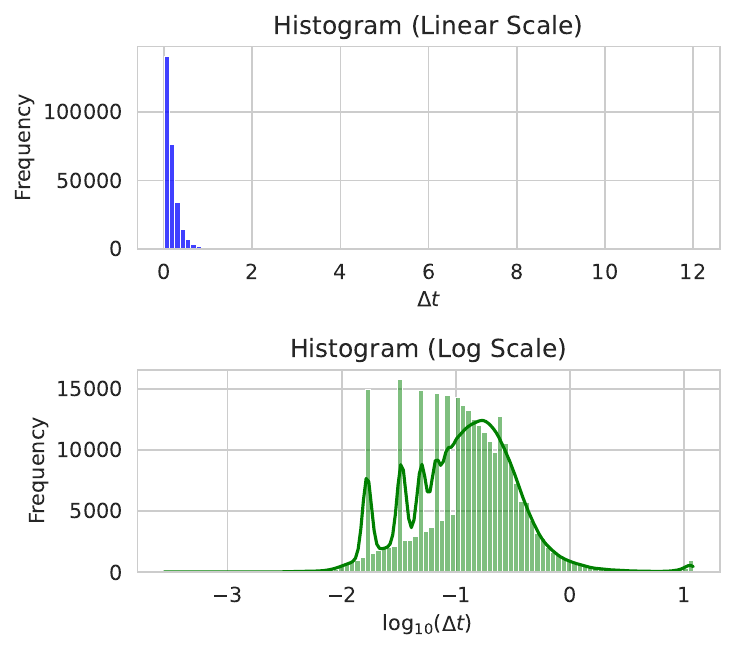}
    \caption{Distribution of relative time intervals ($\Delta t_i$) for the NYC Taxi dataset, showing linear scale (top) and log scale (bottom).}
    \label{fig:dist-taxi}
\end{figure}

\begin{figure}[h!]
    \centering
    \includegraphics[width=.9\columnwidth]{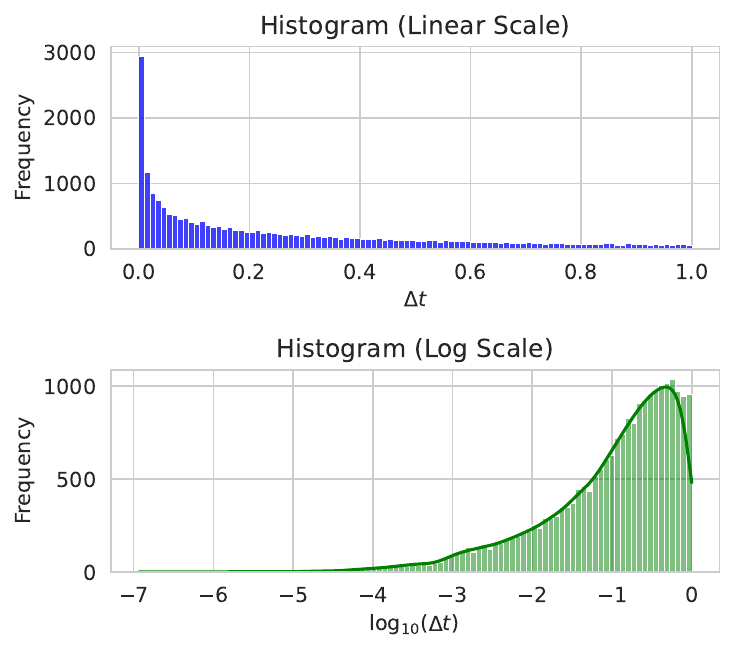}
    \caption{Distribution of relative time intervals ($\Delta t_i$) for the US Earthquake dataset, showing linear scale (top) and log scale (bottom).}
    \label{fig:dist-earthquake}
\end{figure}

\begin{figure}[h!]
    \centering
    \includegraphics[width=.9\columnwidth]{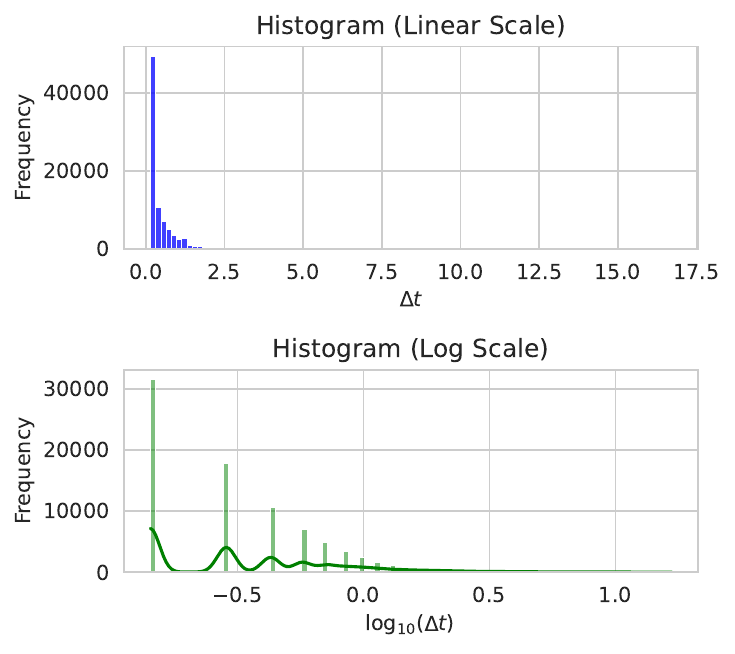}
    \caption{Distribution of relative time intervals ($\Delta t_i$) for the Amazon Review dataset, showing linear scale (top) and log scale (bottom).}
    \label{fig:dist-amazon}
\end{figure}

This appendix provides the full distributions for the relative time intervals ($\Delta t_i$) for all five datasets: Stack Overflow (Figure~\ref{fig:dist-stackoverflow}), Chicago Crime (Figure~\ref{fig:dist-chicago}), NYC Taxi (Figure~\ref{fig:dist-taxi}), US Earthquake (Figure~\ref{fig:dist-earthquake}), and Amazon Review (Figure~\ref{fig:dist-amazon}). Each figure displays the distribution on both a linear scale (top) and a log scale (bottom) to show the overall shape and highlight the behavior of smaller interval values.

%%%%%%%%%%%%%%%%%%%%%%%%%%%%%%%%%%%%%%%%
\section{Data Examples}
\label{app:data-examples}

This appendix provides tokenization examples for a single event sequence from the Stack Overflow training set. For brevity, we display only the first two events using the type-time prompt template.

\textbf{Numeric String Tokenization:} \texttt{'<|begin\_of\_event|>', '<|type\_prefix|>', 'G', 'uru', '<|time\_prefix|>', '0', '.', '000', '000', '<|end\_of\_event|>', '<|begin\_of\_event|>', '<|type\_prefix|>', 'Good', 'ĠAnswer', '<|time\_prefix|>', '0', '.', '437', '604', '<|end\_of\_event|>', ...}

\textbf{Byte Tokenization:} \texttt{'<|begin\_of\_event|>', '<|type\_prefix|>', 'G', 'uru', '<|time\_prefix|>', '<|byte\_000|>', '<|byte\_000|>', '<|byte\_000|>', '<|byte\_000|>', '<|end\_of\_event|>', '<|begin\_of\_event|>', '<|type\_prefix|>', 'Good', 'ĠAnswer', '<|time\_prefix|>', '<|byte\_147|>', '<|byte\_013|>', '<|byte\_224|>', '<|byte\_062|>', '<|end\_of\_event|>', ...}

\textbf{Absolute Calendar Tokenization (Second Resolution):} \texttt{'<|begin\_of\_event|>', '<|type\_prefix|>', 'G', 'uru', '<|time\_prefix|>', '<|year\_2022|>', '<|month\_01|>', '<|day\_13|>', '<|hour\_02|>', '<|min\_52|>', '<|sec\_08|>', '<|end\_of\_event|>', '<|begin\_of\_event|>', '<|type\_prefix|>', 'Good', 'ĠAnswer', '<|time\_prefix|>', '<|year\_2022|>', '<|month\_01|>', '<|day\_26|>', '<|hour\_05|>', '<|min\_56|>', '<|sec\_36|>', '<|end\_of\_event|>', ...}

\textbf{Absolute Calendar Tokenization (Day Resolution):} \texttt{'<|begin\_of\_event|>', '<|type\_prefix|>', 'G', 'uru', '<|time\_prefix|>', '<|year\_2022|>', '<|month\_01|>', '<|day\_13|>', '<|end\_of\_event|>', '<|begin\_of\_event|>', '<|type\_prefix|>', 'Good', 'ĠAnswer', '<|time\_prefix|>', '<|year\_2022|>', '<|month\_01|>', '<|day\_26|>', '<|end\_of\_event|>', ...}

\textbf{Relative Calendar Tokenization (Second Resolution):} \texttt{'<|begin\_of\_event|>', '<|type\_prefix|>', 'G', 'uru', '<|time\_prefix|>', '<|year\_00|>', '<|month\_00|>', '<|day\_00|>', '<|hour\_00|>', '<|min\_00|>', '<|sec\_00|>', '<|end\_of\_event|>', '<|begin\_of\_event|>', '<|type\_prefix|>', 'Good', 'ĠAnswer', '<|time\_prefix|>', '<|year\_00|>', '<|month\_00|>', '<|day\_13|>', '<|hour\_03|>', '<|min\_04|>', '<|sec\_28|>', '<|end\_of\_event|>', ...}

\textbf{Relative Calendar Tokenization (Day Resolution):} \texttt{'<|begin\_of\_event|>', '<|type\_prefix|>', 'G', 'uru', '<|time\_prefix|>', '<|year\_00|>', '<|month\_00|>', '<|day\_00|>', '<|end\_of\_event|>', '<|begin\_of\_event|>', '<|type\_prefix|>', 'Good', 'ĠAnswer', '<|time\_prefix|>', '<|year\_00|>', '<|month\_00|>', '<|day\_13|>', '<|end\_of\_event|>', ...}

\textbf{Scale Bin Tokenization (Linear Scale):} \texttt{'<|begin\_of\_event|>', '<|type\_prefix|>', 'G', 'uru', '<|time\_prefix|>', '<|bin\_000|>', '<|end\_of\_event|>', '<|begin\_of\_event|>', '<|type\_prefix|>', 'Good', 'ĠAnswer', '<|time\_prefix|>', '<|bin\_013|>', '<|end\_of\_event|>', ...}

\textbf{Scale Bin Tokenization (Log Scale):} \texttt{'<|begin\_of\_event|>', '<|type\_prefix|>', 'G', 'uru', '<|time\_prefix|>', '<|bin\_000|>', '<|end\_of\_event|>', '<|begin\_of\_event|>', '<|type\_prefix|>', 'Good', 'ĠAnswer', '<|time\_prefix|>', '<|bin\_189|>', '<|end\_of\_event|>', ...}

\textbf{Residual Scalar Quantization Tokenization (Linear Scale, 1 Level):} \texttt{'<|begin\_of\_event|>', '<|type\_prefix|>', 'G', 'uru', '<|time\_prefix|>', '<|L0\_059|>', '<|end\_of\_event|>', '<|begin\_of\_event|>', '<|type\_prefix|>', 'Good', 'ĠAnswer', '<|time\_prefix|>', '<|L0\_047|>', '<|end\_of\_event|>', ...}

\textbf{Residual Scalar Quantization Tokenization (Linear Scale, 4 Levels):} \texttt{'<|begin\_of\_event|>', '<|type\_prefix|>', 'G', 'uru', '<|time\_prefix|>', '<|L0\_10|>', '<|L1\_13|>', '<|L2\_21|>', '<|L3\_46|>', '<|end\_of\_event|>', '<|begin\_of\_event|>', '<|type\_prefix|>', 'Good', 'ĠAnswer', '<|time\_prefix|>', '<|L0\_42|>', '<|L1\_42|>', '<|L2\_44|>', '<|L3\_15|>', '<|end\_of\_event|>', ...}

\textbf{Residual Scalar Quantization Tokenization (Log Scale, 1 Level):} \texttt{'<|begin\_of\_event|>', '<|type\_prefix|>', 'G', 'uru', '<|time\_prefix|>', '<|L0\_238|>', '<|end\_of\_event|>', '<|begin\_of\_event|>', '<|type\_prefix|>', 'Good', 'ĠAnswer', '<|time\_prefix|>', '<|L0\_118|>', '<|end\_of\_event|>', ...}

\textbf{Residual Scalar Quantization Tokenization (Log Scale, 4 Levels):} \texttt{'<|begin\_of\_event|>', '<|type\_prefix|>', 'G', 'uru', '<|time\_prefix|>', '<|L0\_38|>', '<|L1\_35|>', '<|L2\_22|>', '<|L3\_28|>', '<|end\_of\_event|>', '<|begin\_of\_event|>', '<|type\_prefix|>', 'Good', 'ĠAnswer', '<|time\_prefix|>', '<|L0\_41|>', '<|L1\_45|>', '<|L2\_43|>', '<|L3\_57|>', '<|end\_of\_event|>', ...}

%%%%%%%%%%%%%%%%%%%%%%%%%%%%%%%%%%%%%%%%
\section{Experimental Settings}
\label{app:experimental-settings}

In this research, we fine-tune a pretrained LLM backbone Llama-3.2-1B \citep{grattafiori2024llama} for all experiments, employing 4-bit quantization \citep{dettmers2023qlora} for efficiency. We use Parameter-Efficient Fine-Tuning (PEFT) via Low-Rank Adaptation (LoRA) \citep{hu2022lora} with a rank $r=16$ and $\alpha=32$, applying updates to all attention projection matrices. All models are trained for 5 epochs using the AdamW optimizer \citep{loshchilov2019decoupled} with a cosine learning rate scheduler, a learning rate of $0.001$, and a warmup ratio of $0.1$ through the Hugging Face framework \citep{wolf2020transformers}. We use a per-device train batch size of 4 with 4 gradient accumulation steps, resulting in an effective batch size of 16. All hyper-parameters were determined through preliminary experiments and fixed for the main experiments to avoid exhaustive tuning. 

When using tokenization strategies that expand the LLM's vocabulary with new temporal tokens, we explicitly set their indices to be trainable within the LoRA configuration to ensure their initialized embeddings are fully optimized alongside the adapters. For the residual scalar quantization (RSQ) tokenizer, we utilize K-Means and default parameter settings for initialization, convergence, and optimization from scikit-learn \citep{pedregosa2011scikit}. We use TPP-LLM \citep{liu2024tpp} as our baseline, configured with the same foundation model and QLoRA settings. We follow the original TPP-LLM paper for most hyper-parameters; however, to avoid overfitting, we replace their constant learning rate scheduler with a cosine scheduler and train for 10 epochs, using the same warmup ratio as our experiments. All experiments were conducted on NVIDIA A100 GPUs.

%%%%%%%%%%%%%%%%%%%%%%%%%%%%%%%%%%%%%%%%
\section{Ablation Studies}
\label{app:ablation-studies}

We further investigate the impact of key architectural and design choices through ablation studies on quantization granularity, temporal resolution, prompt template ordering, and the scaling of both the foundation model and the dataset. For these analyses, we selected two representative datasets: Stack Overflow, representing a skewed log-normal distribution, and NYC Taxi, representing a mixed distribution with both log-normal and spiky characteristics.

%%%%%%%%%%%%%%%%%%%%%%%%%%%%%%%%%%%%%%%%
\subsection{Quantization Levels}

We analyze the impact of quantization granularity by distributing a fixed budget of 256 newly added tokens across different numbers of RSQ levels, as detailed in Table~\ref{tab:ablation-rsq-levels}. Consistent with our main results, the choice of base scale (linear vs. log) remains the dominant factor determining performance. However, within the optimal scale for each dataset, we observe a consistent trend where increasing the number of levels improves predictive precision if a suitable scale is selected. For both Stack Overflow (in log scale) and NYC Taxi (in linear scale), the multi-level configurations achieve lower RMSEs compared to the single-level baselines. This demonstrates that a compositional, hierarchical representation, where tokens iteratively refine the residual error, provides superior temporal precision compared to a flat vocabulary of the same size.

\begin{table}[h!]
\centering
\caption{Ablation study on RSQ levels, comparing strategies that all use a fixed budget of 256 new tokens. Next event type accuracy (Acc \% $\uparrow$) and next event time RMSE ($\downarrow$) are reported, with standard deviations provided as subscripts.}
\label{tab:ablation-rsq-levels}
\resizebox{\linewidth}{!}{%
\setlength{\tabcolsep}{3pt}
\begin{tabular}{l l rr rr}
\toprule
 & & \multicolumn{2}{c}{\textbf{Stack Overflow}} & \multicolumn{2}{c}{\textbf{NYC Taxi}} \\
\cmidrule(lr){3-4} \cmidrule(lr){5-6}
\textbf{Tokenizer} & \textbf{Bins} & \textbf{Acc $\uparrow$} & \textbf{RMSE $\downarrow$} & \textbf{Acc $\uparrow$} & \textbf{RMSE $\downarrow$} \\
\midrule
% --- Linear Scale Baselines ---
RSQ (Linear) & 256 & 44.2$_{\pm.1}$ & 0.595$_{\pm.004}$ & \textbf{92.0}$_{\pm.0}$ & 0.911$_{\pm.017}$ \\
RSQ (Linear) & 128-128 & 44.3$_{\pm.1}$ & 0.602$_{\pm.001}$ & \textbf{92.0}$_{\pm.0}$ & 0.921$_{\pm.003}$ \\
RSQ (Linear) & 85-85-86 & \underline{44.4}$_{\pm.1}$ & 0.591$_{\pm.004}$ & \textbf{92.0}$_{\pm.0}$ & \underline{0.904}$_{\pm.004}$ \\
RSQ (Linear) & 64-64-64-64 & \textbf{44.5}$_{\pm.0}$ & 0.603$_{\pm.002}$ & \textbf{92.0}$_{\pm.0}$ & \textbf{0.891}$_{\pm.007}$ \\
\midrule
% --- Log Scale Baselines ---
RSQ (Log) & 256 & 44.3$_{\pm.1}$ & 0.497$_{\pm.005}$ & \textbf{92.0}$_{\pm.0}$ & 1.221$_{\pm.040}$ \\
RSQ (Log) & 128-128 & 44.3$_{\pm.1}$ & \underline{0.480}$_{\pm.002}$ & \textbf{92.0}$_{\pm.0}$ & 1.191$_{\pm.041}$ \\
RSQ (Log) & 85-85-86 & 44.3$_{\pm.0}$ & 0.494$_{\pm.002}$ & \textbf{92.0}$_{\pm.0}$ & 1.091$_{\pm.020}$ \\
RSQ (Log) & 64-64-64-64 & 44.2$_{\pm.1}$ & \textbf{0.474}$_{\pm.003}$ & \textbf{92.0}$_{\pm.0}$ & 1.201$_{\pm.057}$ \\
\bottomrule
\end{tabular}
}
\end{table}

%%%%%%%%%%%%%%%%%%%%%%%%%%%%%%%%%%%%%%%%
\subsection{Temporal Resolutions}

We investigate the impact of tokenizer granularity by varying the resolution of calendar-based strategies from day down to second, as shown in Table~\ref{tab:ablation-resolution}. The results highlight a critical dependency between the tokenizer's temporal resolution and the intrinsic time scale of the dataset. For high-frequency domains like NYC Taxi, where events occur within minutes, a coarse day resolution is insufficient, leading to catastrophically high RMSE scores for the absolute calendar strategy. Increasing the resolution to minute or second drastically corrects this, recovering performance. In contrast, for the Stack Overflow dataset, which operates on a longer time scale, the performance gains from finer granularity are marginal, with the minute resolution offering a slight optimal balance before plateauing. This confirms that while higher resolution generally preserves more information, the optimal choice is dictated by the characteristic granularity of the event sequence itself.

\begin{table}[h!]
\centering
\caption{Ablation study on temporal resolution. We compare the performance of calendar tokenizers across varying levels of granularity, from day to second. Next event type accuracy (Acc \% $\uparrow$) and next event time RMSE ($\downarrow$) are reported, with standard deviations provided as subscripts.}
\label{tab:ablation-resolution}
\resizebox{\linewidth}{!}{%
\setlength{\tabcolsep}{3pt}
\begin{tabular}{l l rr rr}
\toprule
 & & \multicolumn{2}{c}{\textbf{Stack Overflow}} & \multicolumn{2}{c}{\textbf{NYC Taxi}} \\
\cmidrule(lr){3-4} \cmidrule(lr){5-6}
\textbf{Tokenizer} & \textbf{Resolution} & \textbf{Acc $\uparrow$} & \textbf{RMSE $\downarrow$} & \textbf{Acc $\uparrow$} & \textbf{RMSE $\downarrow$} \\
\midrule
% --- Abs. Calendar ---
Abs. Calendar & Day & 44.7$_{\pm.1}$ & 0.578$_{\pm.002}$ & 91.9$_{\pm.1}$ & 16.086$_{\pm.009}$ \\
Abs. Calendar & Hour & \underline{44.8}$_{\pm.2}$ & \underline{0.568}$_{\pm.003}$ & \textbf{92.2}$_{\pm.0}$ & 1.132$_{\pm.005}$ \\
Abs. Calendar & Minute & \textbf{44.9}$_{\pm.1}$ & \textbf{0.566}$_{\pm.002}$ & \textbf{92.2}$_{\pm.0}$ & \underline{0.873}$_{\pm.003}$ \\
Abs. Calendar & Second & \underline{44.8}$_{\pm.1}$ & 0.569$_{\pm.003}$ & \textbf{92.2}$_{\pm.0}$ & \textbf{0.869}$_{\pm.006}$ \\
\midrule
% --- Rel. Calendar ---
Rel. Calendar & Day & 44.4$_{\pm.1}$ & 0.626$_{\pm.000}$ & 91.9$_{\pm.0}$ & 0.912$_{\pm.000}$ \\
Rel. Calendar & Hour & 44.3$_{\pm.1}$ & 0.624$_{\pm.001}$ & 91.9$_{\pm.1}$ & 0.912$_{\pm.001}$ \\
Rel. Calendar & Minute & 44.3$_{\pm.1}$ & 0.624$_{\pm.001}$ & \underline{92.0}$_{\pm.0}$ & 0.892$_{\pm.001}$ \\
Rel. Calendar & Second & 44.5$_{\pm.1}$ & 0.623$_{\pm.001}$ & \underline{92.0}$_{\pm.0}$ & 0.891$_{\pm.002}$ \\
\bottomrule
\end{tabular}
}
\end{table}

%%%%%%%%%%%%%%%%%%%%%%%%%%%%%%%%%%%%%%%%
\subsection{Template Orders}

We explore the sensitivity of model performance to the ordering of information within the prompt template, specifically, whether to place event type tokens before or after event time tokens. As shown in Table~\ref{tab:ablation-order}, our results consistently favor the ``Type-Time'' ordering (placing the event type before the event time tokens). Notably, on the Stack Overflow dataset, reversing this order to ``Time-Type'' causes a sharp degradation in event type prediction accuracy for strategies that generate long or semantically opaque sequences, such as numeric string, byte, and relative calendar tokenization. This suggests that intervening complex temporal sequences between event types may dilute the semantic context required for accurate type prediction. Furthermore, in several cases (e.g., RSQ on NYC Taxi), the ``Time-Type'' order also negatively impacts temporal RMSE. Based on these observations, we standardize on the ``Type-Time'' template for all main experiments to ensure optimal performance.

\begin{table}[h!]
\centering
\caption{Ablation study on template order for testing the effect of placing event type tokens or event time tokens first in the sequence. Next event type accuracy (Acc \% $\uparrow$) and next event time RMSE ($\downarrow$) are reported, with standard deviations provided as subscripts.}
\label{tab:ablation-order}
\resizebox{\linewidth}{!}{%
\setlength{\tabcolsep}{3pt}
\begin{tabular}{l l rr rr}
\toprule
 & & \multicolumn{2}{c}{\textbf{Stack Overflow}} & \multicolumn{2}{c}{\textbf{NYC Taxi}} \\
\cmidrule(lr){3-4} \cmidrule(lr){5-6}
\textbf{Tokenizer} & \textbf{Template} & \textbf{Acc $\uparrow$} & \textbf{RMSE $\downarrow$} & \textbf{Acc $\uparrow$} & \textbf{RMSE $\downarrow$} \\
\midrule
% --- Numeric String ---
Numeric String & Type-Time & \underline{44.5}$_{\pm.1}$ & 0.625$_{\pm.000}$ & \underline{92.0}$_{\pm.0}$ & 0.888$_{\pm.002}$ \\
Numeric String & Time-Type & 33.3$_{\pm3.0}$ & 0.625$_{\pm.000}$ & 91.8$_{\pm.0}$ & 0.892$_{\pm.001}$ \\
\midrule
% --- Byte ---
Byte & Type-Time & 44.3$_{\pm.0}$ & 0.516$_{\pm.004}$ & \underline{92.0}$_{\pm.0}$ & 0.892$_{\pm.001}$ \\
Byte & Time-Type & 39.0$_{\pm.9}$ & 0.540$_{\pm.004}$ & 91.7$_{\pm.0}$ & 0.894$_{\pm.000}$ \\
\midrule
% --- Abs. Calendar ---
Abs. Calendar (Sec.) & Type-Time & \textbf{44.8}$_{\pm.1}$ & 0.569$_{\pm.003}$ & \textbf{92.2}$_{\pm.0}$ & \textbf{0.869}$_{\pm.006}$ \\
Abs. Calendar (Sec.) & Time-Type & 43.6$_{\pm.4}$ & 3.835$_{\pm2.771}$ & 91.9$_{\pm.0}$ & \underline{0.874}$_{\pm.006}$ \\
\midrule
% --- Rel. Calendar ---
Rel. Calendar (Sec.) & Type-Time & \underline{44.5}$_{\pm.1}$ & 0.623$_{\pm.001}$ & \underline{92.0}$_{\pm.0}$ & 0.891$_{\pm.002}$ \\
Rel. Calendar (Sec.) & Time-Type & 39.0$_{\pm2.0}$ & 0.623$_{\pm.000}$ & 91.8$_{\pm.1}$ & 0.894$_{\pm.004}$ \\
\midrule
% --- Scale Bin (Linear) ---
Scale Bin (Linear) & Type-Time & 44.3$_{\pm.1}$ & 0.615$_{\pm.000}$ & \underline{92.0}$_{\pm.0}$ & 0.894$_{\pm.007}$ \\
Scale Bin (Linear) & Time-Type & 43.8$_{\pm.1}$ & 0.614$_{\pm.001}$ & 91.9$_{\pm.0}$ & 0.906$_{\pm.007}$ \\
\midrule
% --- Scale Bin (Log) ---
Scale Bin (Log) & Type-Time & 44.3$_{\pm.0}$ & 0.479$_{\pm.003}$ & \underline{92.0}$_{\pm.0}$ & 1.127$_{\pm.061}$ \\
Scale Bin (Log) & Time-Type & 44.1$_{\pm.1}$ & \underline{0.478}$_{\pm.003}$ & 91.7$_{\pm.0}$ & 1.312$_{\pm.016}$ \\
\midrule
% --- RSQ (Linear, L4) ---
RSQ (Linear, L4) & Type-Time & 44.5$_{\pm.0}$ & 0.603$_{\pm.002}$ & \underline{92.0}$_{\pm.0}$ & 0.891$_{\pm.007}$ \\
RSQ (Linear, L4) & Time-Type & 43.7$_{\pm.1}$ & 0.605$_{\pm.001}$ & 91.9$_{\pm.0}$ & 0.898$_{\pm.015}$ \\
\midrule
% --- RSQ (Log, L4) ---
RSQ (Log, L4) & Type-Time & 44.2$_{\pm.1}$ & \textbf{0.474}$_{\pm.003}$ & \underline{92.0}$_{\pm.0}$ & 1.201$_{\pm.057}$ \\
RSQ (Log, L4) & Time-Type & 44.1$_{\pm.1}$ & \textbf{0.474}$_{\pm.002}$ & 91.7$_{\pm.1}$ & 1.479$_{\pm.104}$ \\
\bottomrule
\end{tabular}
}
\end{table}

%%%%%%%%%%%%%%%%%%%%%%%%%%%%%%%%%%%%%%%%
\subsection{Model Sizes}

To evaluate the impact of foundation model scale, we compare the performance of our Llama-3.2-1B backbone against the larger Llama-3.2-3B, with results presented in Table~\ref{tab:ablation-model-size}. The findings show that under our current fine-tuning setting and data size, the 1B model already achieves a highly competitive performance, establishing a strong and efficient baseline. The 3B model performs comparably, and in some cases, such as with the absolute calendar tokenizer on Stack Overflow, achieves a slight improvement in both accuracy and RMSE. The consistent performance trends observed between these two scales suggest that the fundamental trade-offs are rooted in the tokenization logic itself rather than specific model capacity, providing strong confidence in their generalizability to larger architectures. Thus, while the Llama-3.2-1B model provides an excellent balance of performance and efficiency for this task setup, the framework remains fully compatible with larger models, which may yield further advantages with more extensive training data.

\begin{table}[h!]
\centering
\caption{Ablation study on foundation model size, comparing Llama-3.2-1B and Llama-3.2-3B backbones. Next event type accuracy (Acc \% $\uparrow$) and next event time RMSE ($\downarrow$) are reported, with standard deviations provided as subscripts.}
\label{tab:ablation-model-size}
\resizebox{\linewidth}{!}{%
\setlength{\tabcolsep}{3pt}
\begin{tabular}{l l rr rr}
\toprule
 & & \multicolumn{2}{c}{\textbf{Stack Overflow}} & \multicolumn{2}{c}{\textbf{NYC Taxi}} \\
\cmidrule(lr){3-4} \cmidrule(lr){5-6}
\textbf{Tokenizer} & \textbf{Model} & \textbf{Acc $\uparrow$} & \textbf{RMSE $\downarrow$} & \textbf{Acc $\uparrow$} & \textbf{RMSE $\downarrow$} \\
\midrule
% --- Numeric String ---
Numeric String & 1B & 44.5$_{\pm.1}$ & 0.625$_{\pm.000}$ & 92.0$_{\pm.0}$ & 0.888$_{\pm.002}$ \\
Numeric String & 3B & 43.7$_{\pm1.2}$ & 0.625$_{\pm.001}$ & \underline{92.1}$_{\pm.0}$ & 0.897$_{\pm.001}$ \\
\midrule
% --- Byte ---
Byte & 1B & 44.3$_{\pm.0}$ & 0.516$_{\pm.004}$ & 92.0$_{\pm.0}$ & 0.892$_{\pm.001}$ \\
Byte & 3B & 43.7$_{\pm.6}$ & 0.552$_{\pm.047}$ & 92.0$_{\pm.0}$ & 0.893$_{\pm.002}$ \\
\midrule
% --- Abs. Calendar ---
Abs. Calendar (Sec.) & 1B & \underline{44.8}$_{\pm.1}$ & 0.569$_{\pm.003}$ & \textbf{92.2}$_{\pm.0}$ & \textbf{0.869}$_{\pm.006}$ \\
Abs. Calendar (Sec.) & 3B & \textbf{45.0}$_{\pm.1}$ & 0.564$_{\pm.000}$ & \textbf{92.2}$_{\pm.1}$ & \underline{0.870}$_{\pm.005}$ \\
\midrule
% --- Rel. Calendar ---
Rel. Calendar (Sec.) & 1B & 44.5$_{\pm.1}$ & 0.623$_{\pm.001}$ & 92.0$_{\pm.0}$ & 0.891$_{\pm.002}$ \\
Rel. Calendar (Sec.) & 3B & 44.3$_{\pm.1}$ & 0.625$_{\pm.000}$ & 92.0$_{\pm.0}$ & 0.895$_{\pm.006}$ \\
\midrule
% --- Scale Bin (Linear) ---
Scale Bin (Linear) & 1B & 44.3$_{\pm.1}$ & 0.615$_{\pm.000}$ & 92.0$_{\pm.0}$ & 0.894$_{\pm.007}$ \\
Scale Bin (Linear) & 3B & 44.0$_{\pm.4}$ & 0.614$_{\pm.001}$ & \underline{92.1}$_{\pm.0}$ & 0.908$_{\pm.012}$ \\
\midrule
% --- Scale Bin (Log) ---
Scale Bin (Log) & 1B & 44.3$_{\pm.0}$ & 0.479$_{\pm.003}$ & 92.0$_{\pm.0}$ & 1.127$_{\pm.061}$ \\
Scale Bin (Log) & 3B & 44.1$_{\pm.2}$ & 0.481$_{\pm.002}$ & 92.0$_{\pm.0}$ & 1.161$_{\pm.060}$ \\
\midrule
% --- RSQ (Linear, L4) ---
RSQ (Linear, L4) & 1B & 44.5$_{\pm.0}$ & 0.603$_{\pm.002}$ & 92.0$_{\pm.0}$ & 0.891$_{\pm.007}$ \\
RSQ (Linear, L4) & 3B & 44.4$_{\pm.1}$ & 0.602$_{\pm.001}$ & 92.0$_{\pm.0}$ & 0.906$_{\pm.004}$ \\
\midrule
% --- RSQ (Log, L4) ---
RSQ (Log, L4) & 1B & 44.2$_{\pm.1}$ & \underline{0.474}$_{\pm.003}$ & 92.0$_{\pm.0}$ & 1.201$_{\pm.057}$ \\
RSQ (Log, L4) & 3B & 44.2$_{\pm.1}$ & \textbf{0.473}$_{\pm.001}$ & 92.0$_{\pm.0}$ & 1.294$_{\pm.069}$ \\
\bottomrule
\end{tabular}
}
\end{table}

%%%%%%%%%%%%%%%%%%%%%%%%%%%%%%%%%%%%%%%%
\subsection{Data Sizes}

To evaluate the impact of dataset scale on performance, we extend the Stack Overflow dataset from 2 years (S) to 2.5 years (M) and 3 years (L), and the NYC Taxi dataset from 7 days (S) to 10 days (M) and 14 days (L), as detailed in Table~\ref{tab:larger-dataset-stats}. To ensure a fair comparison, all evaluations are performed on the test set of the smallest data variant (S). The results in Table~\ref{tab:ablation-data-scale} show mixed outcomes. For NYC Taxi, increasing the data size generally yields slight improvements in accuracy and RMSE across most tokenizers, with the Absolute Calendar (Sec.) achieving the best results on the largest split. However, for Stack Overflow, performance gains are inconsistent or saturate; while some tokenizers like Scale Bin (Log) improve with more data, others plateau or slightly degrade. This suggests that the capacity of our LoRA-based fine-tuning may be saturating as the dataset complexity grows. While parameter-efficient fine-tuning (PEFT) is effective for smaller datasets, scaling to significantly larger event sequences may require full fine-tuning or continual pre-training (CPT) to fully leverage the additional data, a promising direction for future work.

\begin{table}[h!]
\centering
\caption{Statistics for the data scale ablation study. We compare small (S), medium (M), and large (L) variants of the Stack Overflow and NYC Taxi datasets, detailing event types, total events, sequence with their train/validation/test splits, average sequence length, and temporal units.}
\label{tab:larger-dataset-stats}
\resizebox{\linewidth}{!}{%
\begin{tabular}{llrrrrrl}
\toprule
\textbf{Dataset} & \textbf{Size} & \textbf{Types} & \textbf{Events} & \textbf{Seqs} & \textbf{Train/Val/Test} & \textbf{Seq Len} & \textbf{Unit} \\
\midrule
Stack Overflow & S & 25 & 187,836 & 3,336 & 2,668/334/334 & 56.31 & Month \\
Stack Overflow & M & 27 & 308,287 & 5,460 & 4,368/546/546 & 56.46 & Month \\
Stack Overflow & L & 27 & 459,269 & 8,065 & 6,452/806/807 & 56.90 & Month \\
NYC Taxi & S & 8 & 362,370 & 2,957 & 2,365/296/296 & 122.55 & Hour \\
NYC Taxi & M & 8 & 444,170 & 3,641 & 2,912/364/365 & 121.99 & Hour \\
NYC Taxi & L & 8 & 680,510 & 5,579 & 4,463/558/558 & 121.98 & Hour \\
\bottomrule
\end{tabular}
}
\end{table}

\begin{table}[h!]
\centering
\caption{Ablation study on data scale. We compare the performance of tokenization strategies across three data scale variants (small, medium, and large) on datasets. Next event type accuracy (Acc \% $\uparrow$) and next event time RMSE ($\downarrow$) are reported, with standard deviations provided as subscripts.}
\label{tab:ablation-data-scale}
\resizebox{\linewidth}{!}{%
\setlength{\tabcolsep}{3pt}
\begin{tabular}{l l cc cc}
\toprule
 & & \multicolumn{2}{c}{\textbf{Stack Overflow}} & \multicolumn{2}{c}{\textbf{NYC Taxi}} \\
\cmidrule(lr){3-4} \cmidrule(lr){5-6}
\textbf{Tokenizer} & \textbf{Data} & \textbf{Acc $\uparrow$} & \textbf{RMSE $\downarrow$} & \textbf{Acc $\uparrow$} & \textbf{RMSE $\downarrow$} \\
\midrule
% -- Numeric String --
Numeric String & S & 44.5$_{\pm.1}$ & 0.625$_{\pm.000}$ & 92.0$_{\pm.0}$ & 0.888$_{\pm.002}$ \\
 & M & \textbf{45.0}$_{\pm.1}$ & 0.625$_{\pm.000}$ & \underline{92.2}$_{\pm.0}$ & 0.888$_{\pm.001}$ \\
 & L & 44.4$_{\pm.5}$ & 0.625$_{\pm.001}$ & \underline{92.2}$_{\pm.0}$ & 0.887$_{\pm.005}$ \\
\midrule
% -- Byte --
Byte & S & 44.3$_{\pm.0}$ & 0.516$_{\pm.004}$ & 92.0$_{\pm.0}$ & 0.892$_{\pm.001}$ \\
 & M & 44.4$_{\pm.0}$ & 0.514$_{\pm.003}$ & 92.1$_{\pm.0}$ & 0.892$_{\pm.002}$ \\
 & L & 44.0$_{\pm.7}$ & 0.556$_{\pm.052}$ & \underline{92.2}$_{\pm.0}$ & 0.889$_{\pm.002}$ \\
\midrule
% -- Abs. Calendar (Sec.) --
Abs. Calendar (Sec.) & S & 44.8$_{\pm.1}$ & 0.569$_{\pm.003}$ & \underline{92.2}$_{\pm.0}$ & 0.869$_{\pm.006}$ \\
 & M & 44.6$_{\pm.2}$ & 1.772$_{\pm2.413}$ & \textbf{92.3}$_{\pm.0}$ & \underline{0.860}$_{\pm.004}$ \\
 & L & \textbf{45.0}$_{\pm.1}$ & 0.566$_{\pm.003}$ & \textbf{92.3}$_{\pm.0}$ & \textbf{0.859}$_{\pm.003}$ \\
\midrule
% -- Rel. Calendar (Sec.) --
Rel. Calendar (Sec.) & S & 44.5$_{\pm.1}$ & 0.623$_{\pm.001}$ & 92.0$_{\pm.0}$ & 0.891$_{\pm.002}$ \\
 & M & 44.7$_{\pm.1}$ & 0.624$_{\pm.001}$ & 92.1$_{\pm.0}$ & 0.887$_{\pm.003}$ \\
 & L & 44.1$_{\pm1.1}$ & 0.624$_{\pm.001}$ & \underline{92.2}$_{\pm.0}$ & 0.885$_{\pm.003}$ \\
\midrule
% -- Scale Bin (Linear) --
Scale Bin (Linear) & S & 44.3$_{\pm.1}$ & 0.615$_{\pm.000}$ & 92.0$_{\pm.0}$ & 0.894$_{\pm.007}$ \\
 & M & 44.7$_{\pm.1}$ & 0.612$_{\pm.000}$ & 92.1$_{\pm.1}$ & 0.887$_{\pm.004}$ \\
 & L & 44.4$_{\pm.4}$ & 0.598$_{\pm.000}$ & 92.0$_{\pm.2}$ & 0.896$_{\pm.006}$ \\
\midrule
% -- Scale Bin (Log) --
Scale Bin (Log) & S & 44.3$_{\pm.0}$ & 0.479$_{\pm.003}$ & 92.0$_{\pm.0}$ & 1.127$_{\pm.061}$ \\
 & M & 44.6$_{\pm.2}$ & 0.481$_{\pm.001}$ & \underline{92.2}$_{\pm.0}$ & 1.231$_{\pm.027}$ \\
 & L & \underline{44.9}$_{\pm.1}$ & \underline{0.477}$_{\pm.003}$ & \underline{92.2}$_{\pm.1}$ & 1.169$_{\pm.035}$ \\
\midrule
% -- RSQ (Linear, L4) --
RSQ (Linear, L4) & S & 44.5$_{\pm.0}$ & 0.603$_{\pm.002}$ & 92.0$_{\pm.0}$ & 0.891$_{\pm.007}$ \\
 & M & \textbf{45.0}$_{\pm.1}$ & 0.599$_{\pm.002}$ & \underline{92.2}$_{\pm.0}$ & 0.897$_{\pm.004}$ \\
 & L & 44.8$_{\pm.1}$ & 0.605$_{\pm.004}$ & \underline{92.2}$_{\pm.0}$ & 0.910$_{\pm.008}$ \\
\midrule
% -- RSQ (Log, L4) --
RSQ (Log, L4) & S & 44.2$_{\pm.1}$ & \textbf{0.474}$_{\pm.003}$ & 92.0$_{\pm.0}$ & 1.201$_{\pm.057}$ \\
 & M & 44.7$_{\pm.1}$ & 0.518$_{\pm.004}$ & \underline{92.2}$_{\pm.0}$ & 1.293$_{\pm.054}$ \\
 & L & 44.7$_{\pm.3}$ & 0.512$_{\pm.009}$ & \underline{92.2}$_{\pm.0}$ & 1.158$_{\pm.015}$ \\
\bottomrule
\end{tabular}
}
\end{table}

%%%%%%%%%%%%%%%%%%%%%%%%%%%%%%%%%%%%%%%%
\end{document}